\documentclass[letterpaper, 10 pt, conference]{ieeeconf}
\IEEEoverridecommandlockouts
\usepackage{cite}
\usepackage{booktabs} %
\usepackage{cite}
\usepackage{amsmath,amssymb,amsfonts}
\usepackage{algorithm}
\usepackage{algpseudocode}
\usepackage{graphicx}
\usepackage{textcomp}
\usepackage{xcolor}

\usepackage{amssymb}
\usepackage{bbold}
\usepackage{booktabs}
\usepackage{tabularx}
\usepackage{multirow}
\usepackage{amsfonts}
\usepackage{adjustbox}
\usepackage{lscape}
\usepackage{subcaption}

\def\BibTeX{{\rm B\kern-.05em{\sc i\kern-.025em b}\kern-.08em
    T\kern-.1667em\lower.7ex\hbox{E}\kern-.125emX}}

\usepackage{tikz}
\usetikzlibrary{shapes.geometric, arrows}

\tikzstyle{startstop} = [rectangle, rounded corners, 
minimum width=3cm, 
minimum height=1cm,
text centered, 
draw=black, 
fill=yellow!10]

\tikzstyle{io} = [trapezium, 
trapezium stretches=true, % A later addition
trapezium left angle=70, 
trapezium right angle=110, 
minimum width=3cm, 
minimum height=1cm, text centered, 
draw=black, fill=green!10]

\tikzstyle{process} = [rectangle, 
minimum width=3cm, 
minimum height=1cm, 
text centered, 
text width=3cm, 
draw=black, 
fill=orange!10]

\tikzstyle{decision} = [diamond, 
minimum width=3cm, 
minimum height=1cm, 
text centered, 
draw=black, 
fill=blue!10]
\tikzstyle{arrow} = [thick,->,>=stealth]

\title{
Fin-QD: A Computational Design Framework for Soft Grippers: Integrating MAP-Elites and High-fidelity FEM
}

 \author{Yue Xie$^{1,\dagger}$, Xing Wang$^{2,\dagger}$, Fumiya Iida$^{1}$, David Howard$^{2}$% <-this % stops a space
\thanks{$^{\dagger}$ Equal contributions as first author} 
\thanks{$^{1}$ Department of Engineering, University of Cambridge, UK
        {\tt\small }}%
\thanks{$^{2}$ Robotics Design and Interaction Group, Data61, CSIRO, Australia
        {\tt\small }}%
}

\begin{document} 

\maketitle

\begin{abstract}

Computational design can excite the full potential of soft robotics that has the drawbacks of being highly nonlinear from material, structure, and contact. Up to date, enthusiastic research interests have been demonstrated for individual soft fingers, but the frame design space (how each soft finger is assembled) remains largely unexplored. Computationally design remains challenging for the finger-based soft gripper to grip across multiple geometrical-distinct object types successfully. Including the design space for the gripper frame can bring huge difficulties for conventional optimisation algorithms and fitness calculation methods due to the exponential growth of high-dimensional design space. 
This work proposes an automated computational design optimisation framework that generates gripper diversity to individually grasp geometrically distinct object types based on a quality-diversity approach. This work first discusses a significantly large design space (28 design parameters) for a finger-based soft gripper, including the rarely-explored design space of finger arrangement that is converted to various configurations to arrange individual soft fingers. Then, a contact-based Finite Element Modelling (FEM) is proposed in SOFA to output high-fidelity grasping data for fitness evaluation and feature measurements. Finally, diverse gripper designs are obtained from the framework while considering features such as the volume and workspace of grippers. This work bridges the gap of computationally exploring the vast design space of finger-based soft grippers while grasping large geometrically distinct object types with a simple control scheme.

\end{abstract}

% \begin{keywords}
% Soft robotics, Computational design, Quality diversity,SOFA simulation,  Fin ray gripper
% \end{keywords}

\section{Introduction}
\begin{figure*}[h]
    \centering
\includegraphics[width=0.8\linewidth]{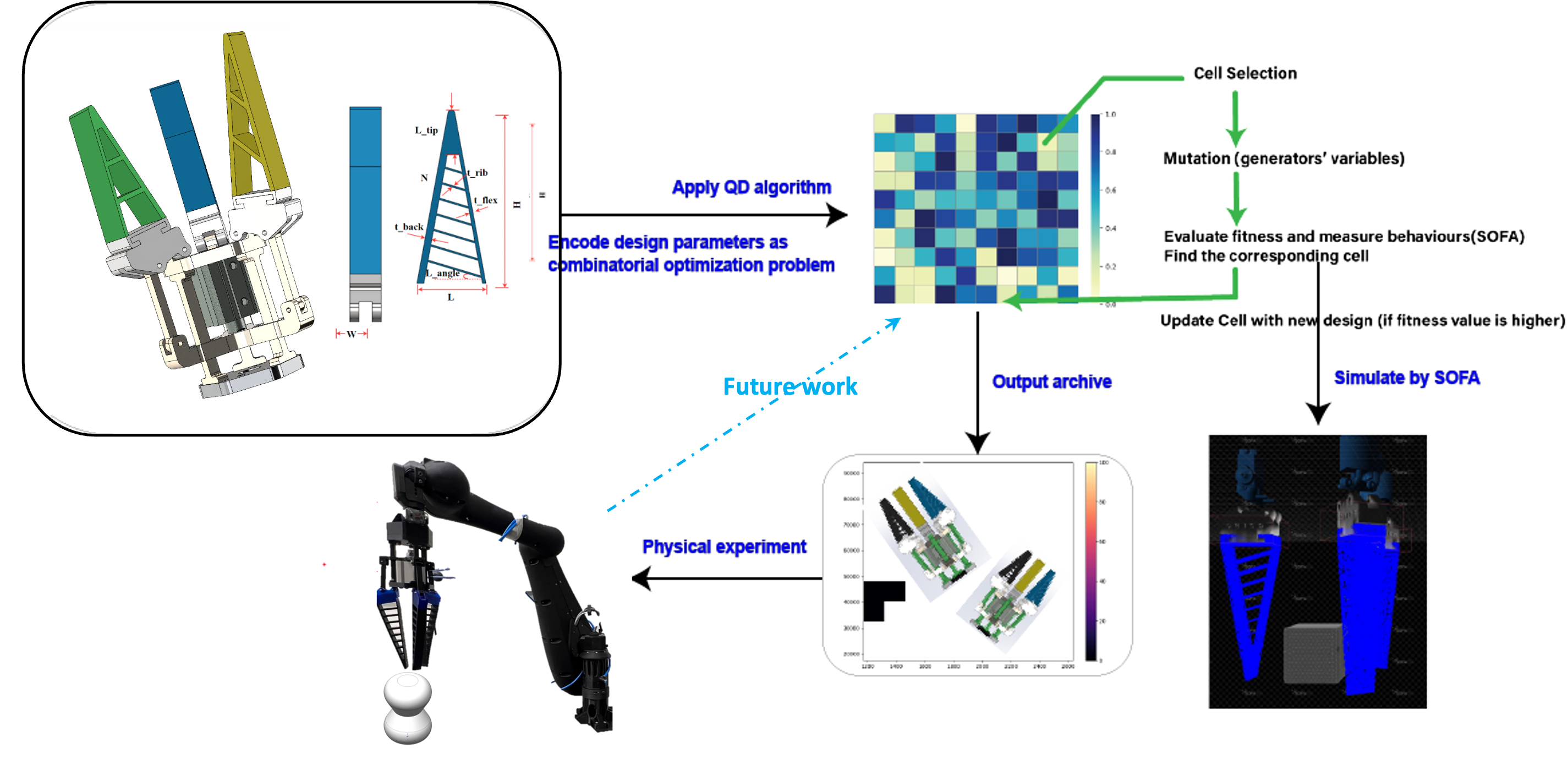}
    \caption{Design optimisation framework to generate diverse gripper designs}
    \label{fig:framework}
\end{figure*}

Recent research in soft robotics~\cite{iida2011soft} has shown the potential benefits of using constitutionally soft materials to design artificial agents. They are frequently designed to mimic biological organisms, such as the soft prosthetic hand~\cite{gu2023soft}, soft robotic arm~\cite{xie2020octopus}, and soft locomotion robot~\cite{chen2021legless}. Benefiting from Soft robotic grippers' effectiveness and popularity in industry applications, such as pick and place in assemble line~\cite{shintake2018soft}, underwater exploration~\cite{aracri2021soft}, and agricultural harvesting~\cite{wang2023development}, designing soft robotic grippers to meet such applications' demands also remains to be of great interest to the research community. 
The design method has gradually moved from human-dominant, bio-inspired approaches to computational design \cite{pinskier2022bioinspiration,howard2019evolving}. 

Computational design shows great potential to output suitable and high-performance soft robotics once given certain design requirements \cite{pinskier2022bioinspiration, howard2022one}. 
It can be initialised by defining a simplified design domain from a human-driven or bio-inspired structure. The behaviour or performance of those initial designs needs to be evaluated using mathematical or physical simulation, which enables the designers to validate or refine the design without performing the physical prototypes.
% It merges the parameterised design, performance evaluation, and optimisation processes within an in silico framework. In detail, it iteratively explores the parameterised design space via frequently utilised computational modelling or relatively rare experimental validation together with the optimisation algorithm.   
Soft robotic gripper benefits significantly from computational design, owing to its intricate nonlinearity from constitutional material, structure design, and contact, and the presence of multimodality in the search space, which pose significant challenges in accurately modelling and searching the design space.

However, to date, computational design has explored a limited design space for finger-based soft grippers, mainly on the design space of individual soft robotic fingers. A significantly larger design space that includes a way to map individual fingers is rarely researched for grasping. 
For example, individuals are mostly assembled on a manually-designed finger mount to form a multiple-finger-based gripper. Liu et al.~\cite{liu2018optimal} generated two optimal finger designs from topology optimisation and manually assembled them symmetrically to form a two-finger gripper. The combination of individuals and their relative positioning has not been properly investigated.   
This unexplored design space is vast and complex as a combination of soft fingers results in an exponential growth of design space. 
% The underlying limitations can be attributed to a lack of an optimisation algorithm that does not degrade for significantly higher dimensional space and a lack of a rapid and high-fidelity fitness model for accurate and fast evaluation of highly nonlinear soft grasping motion. 

Facing this challenge, this study present a computational design framework to investigate the evolution of soft robots while performing contact-based grasping tasks. The work significantly expands the state-of-the-art in the computational design of finger-based soft robotic grippers, involving quality diversity to a general evolutionary system (composed of a rapid and high-fidelity soft robot simulator, a bio-inspired generative encoding, and a quality-diversity EA). 
The vast design space (28 design parameters) is covered and explored by scripting the individual fin ray design and assembling them in certain combinations. A high-fidelity contact-based simulation that covers a complete grasping motion is proposed in the SOFA framework, which outputs the grasping success rate for performance evaluation.   
This work evaluates how the values of different behaviours affect the evolution of different combinations of design parameters and performance in accomplishing tasks. 
Our study also demonstrates the potential of evolutionary techniques to produce a variety of high-performance soft robots, and our analysis may provide new insights into the evolution of soft grippers.

To summarise, the main contributions of this research work are as follows:

    \begin{itemize}    
    \item Propose a highly automated design optimisation framework that generates gripper diversity using a quality-diversity optimization algorithm

    \item Explorer and search for a significantly higher dimensional design space that includes the domain for individual robotic fingers and their combination

    \item Demonstrate a rapid and high-fidelity contact-based FEM to evaluate the complete soft grasping across multiple geometrically distinct objects
    \end{itemize}

The remainder of this paper is organized as follows. Section~\ref{related work} describes the Related work in the research field. Section~\ref{framework} specifies the detailed components for our framework. Then, Section \ref{results} covers the validation and results on our optimised gripper. This study's key conclusions and future directions are summarized in Section~\ref{conclusions}.

\section{Related work}
\label{related work}

Computational design has the potential to create optimal soft robotics by leveraging the power of computation tools despite their challenges of high dimensional design space, nonlinearity and multi-modality search space. It attempts to overcome these challenges by design parameterization, rapid analytical, FEM or physical evaluation, and efficient optimization algorithm \cite{pinskier2022bioinspiration}. 
The genetic optimisation frameworks are of interest here, even though some pioneering frameworks only work for specific types of soft grippers and have restricted the generalisation of the method \cite{mathew2022sorosim}. 

From the optimisation algorithm perspective, traversing the design space involves different levels of computational cost depending on the degree of freedom in the parameterisation. 
Topology optimisation utilises a gradient-based method to design the soft robots by optimising the discrete material distribution under predefined load and boundary constraints \cite{zhang2017design}. The decision variables are normally directly encoded into discrete elements, requiring tens of thousands up to billions of parameters \cite{aage2015topology}. Josh et al.\cite{pinskier2023automated} proposed a pioneering approach to topology optimization that creates soft grippers with multiple materials, and explores techniques for producing hermetically sealed designs. It explores a vast design space with higher simulation accuracy.
The disadvantages of this algorithm include the necessity of detailed loading and boundary conditions for initialisation, the generation of a single feasible design, the significantly large computational cost and less convergence stability. 

Bayesian optimisation (BO) provides a feasible solution to optimise objective functions that are expensive to analyse, resulting in a more efficient traverse of the design space. 
% Bartosz et al.\cite{kaczmarski2023bayesian} proposed a BO to find the optimal fibre-reinforced soft actuator that requires minimum actuation energy under arbitrary control strategy. The optimised design is output from the continuous design space and beats the benchmark design easily. 
Our previous work presented a BO-based computational design framework to search a continuous design parameterisation space \cite{wang2023soro}. A sophisticated FEM based on COMSOL was implemented to find the compliance and normal contact force. It focuses on the design space for individual fingers only and is not applicable for a complete multi-step grasping motion due to the significant expensive computational cost of contact-based FEM.   
Bio-inspired optimisation algorithms demonstrate the capability of solving high dimensional optimisation problem that is challenging for BO that experiences performance degradation. However, because of the iterative, population-based nature of the solver, the computational evolution of soft robots tends to be computationally expensive and relies on low-fidelity solvers rather than high-fidelity FEM for fitness evaluation. This often leads to soft robot designs with notable reality gaps~\cite{pinskier2022bioinspiration}. 

In this work, we aim to promote state-of-the-art of finger-based soft grippers by implementing gradient-free EA, including the quality diversity algorithm to explore the exponential growth design space, the rapid and high-fidelity FEM for grasping, and finally, the generation of soft grippers with diverse features.

\section{Computational Design framework} 
\label{framework}

The section details the working principle of the gradient-free EA framework for soft gripper optimisation, as shown in Figure~\ref{fig:framework}. 
The goal of the computational optimisation is to find the top-ranking gripper designs that demonstrate promising performance in grasping multiple geometrical-distinct objects. 
A benchmark three-fingered gripper is designed with fin ray structure due to its ubiquity throughout research and industry, whose design can be varied by modifying one or any combination of the total 28 design parameters. Once the design is finalised, Gmsh exports high-quality tetrahedron mesh and imports it as TetdrahedronFEMforcefield to the SOFA simulator. The gripper design and configuration are loaded to SOFA for fitness evaluation, which is set to be the successful grasp rate, and is calculated across all target objects to evaluate the performance of the current gripper design. 

% This high-fidelity FEM simulation process considers the complex interaction between all three distinct fingers with the rigid target during contact. The gripper is controlled to grasp under predefined on-off input and is expected to hold the object under gravity. The objective of successful grasp rate is calculated across all target objects to evaluate the performance of the current gripper design. 

\subsection{Problem statement}

The optimisation problem is formulated to discover the mapping between dynamic grasping quality and the parameterised variables that can generate valid designs. 
The grasping quality of a soft robotic gripper can be evaluated by different objectives (fitness functions), such as grasping rate, stability, disturbance resistance, dexterity, etc~\cite{roa2015grasp}. The most intuitive and critical factor is to maximise the grasping success rate, as shown in~\eqref{eqn:1}. It measures the effectiveness of a robot's ability to pick up and hold objects. It is calculated by dividing the number of successful grasps by the total number of attempted grasps. 
The grasping is classified as successful using the Elasped-time threshold, which will be detailed under Section \textit{Rapid simulation}.

    \begin{equation}
        % \[
        \label{eqn:1}
    obj = \arg\max_x \left(f_1(x)\right) \quad  
    \text{s.t.} \quad low_i < x_i < high_i \quad \forall x_i \in x
    % \]
    \end{equation}
While $f_1$ is the objective function (success rate), and $x_i$ represents the design parameters.

\subsection{Gripper Parameterisation}

    \begin{table}[t]
        \caption{Design Variables, symbol, data types and their ranges, $i = \{1,2,3\}$.}
            \centering
          \scalebox{0.75}{
         \makebox[\linewidth][c]{
            \begin{tabular}{lll}
            \toprule
             Variables   &  Symbol & Design range\\
             \midrule
             
             Finger length & $H_{i}$ &  Continuous, $[90, 120]$\\
              Finger width   & $L_{i}$  & Continuous, $[28, 40]$\\
              Finger thickness & $W_{i}$ & Continuous, $[20, 35]$ \\
              Length of solid tip & $L_{tipi}$ & Continuous, $[20, 30]$ \\
              Thickness of the contact surface & $t_{flexi}$ & Continuous, $[1, 3]$  \\
              Thickness of the non-contact surface & $t_{backi}$ & Continuous, $[1, 3]$ \\
              Number of linkage ribs & $N_{i}$ & Discrete, $[1, 10]$ \\
              Thickness of ribs & $t_{ribi}$ & Continuous, $[1, 3]$ \\
              Tilt angle of ribs & $D_{anglei}$ & Continuous, $[-40, 40]$ \\
              
              $\mathbf{G1}$: distance between fingers & $d_{mount}$ & Continuous, $[30, 40]$  \\
              % $\mathbf{G2}$: morphological complexity & $L_{beam}$ & Continuous, $[35, 45]$ \textcolor{red}{remove} \\
            
            \bottomrule
            \label{tab:parameters}
            \end{tabular}}}
        \end{table}

\begin{figure*}[h]
        \centering
        \includegraphics[width=0.6\linewidth]{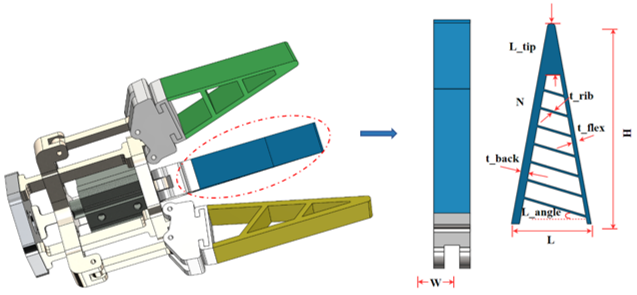}
        \caption{Gripper parameterization: details for the design variables on one fin ray finger}
        \label{fig:param}
    \end{figure*}

The basic gripper design comprises three distinct fin ray fingers, a pneumatic linear actuator for actuation, and rigid finger mounts that link individual fingers. Those rigid finger mounts are programmed with changeable in-plane size $d_{mount}$ to allow a tunable scale of the overall soft gripper. We have added a \textbf{Supplementary Video (SV1)} to demonstrate how the distance between fingers can be modified. 
In addition, nine parameters are assigned for each finger to define its design space, including finger length ($H$), width ($L$), height ($W$), length of solid tip ($L_{tip}$), the thickness of the contact surface ($t_{flex}$), the thickness of the non-contact surface ($t_{back}$), the number of linkage ribs ($N$), the thickness of ribs ($t_{rib}$), and the tilt angle of ribs ($t_{angle}$) (Figure \ref{fig:param}). 
The rich combination of three fingers is coded in the evolutionary algorithm, ending up with 28 design variables (three fingers $\times$ nine variables per finger plus one defining the finger distance). Table ~\ref{tab:parameters} lists the design ranges of all design variables. 
        
Supplementary Video (SV1) is attached to provide a more intuitive explanation of the vast design space of the soft gripper. We replicate the exact same gripper design using the Open Cascade Python library for smooth implementation in the optimisation framework. Note the benchmark gripper has three identical fingers with 9 design parameters being [94.5, 37.5, 21.3, 15, 0, 8, 2, 2, 2] \cite{shan2020modeling} 

\subsection{Rapid simulation}

\begin{figure}[ht]
        \centering
        \includegraphics[width=0.7\linewidth]{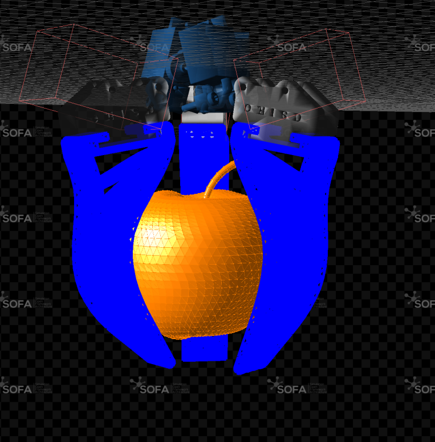}
        \caption{SOFA simulation setup}
        \label{fig:sofa}
    \end{figure}

An open-source physical simulator SOFA for high-fidelity objective calculation is adopted in the framework to evaluate the fitness and measure the behaviours of designed grippers \cite{allard2007sofa}. 
We evaluate the grasping success rate across a set of geometrically distinct objects for each design iteration. Figure \ref{fig:sofa} shows the detailed setup for the high-fidelity SOFA simulation. 
    
SOFA make the simulation a scene with an intrinsic generalised hierarchy. It starts with a parent node, “root”, and comprises more child nodes in a tree structure. Each child node can represent one object (e.g. soft finger) in the simulation scene, and various subnodes of the child node are created to demonstrate the different representations of the same object. For example, the same soft finger may be imported with its visual, mechanical, collision and potentially haptic representations in a simulation scene. 
Our simulation simplifies and accelerates the evaluation by importing three soft fingers and the target objects into the scene. The design space for finger assembly is varied by reloading the relative configuration of all fingers at the beginning of each simulation. A simple controller that generates bending motion for soft robotic fingers is added to the scene. 
The contact is set up between the fingers with the target object using friction contact constraint. Self-collision is enabled for soft fingers to avoid self-penetration on the internal rib region.  

The grasping simulation In SOFA is achieved by discretising the temporal domain into small time steps. The ODESolver is used to construct a linear matrix of the system for further solving. We implemented the EulerImplicitSolver as the timeIntegrationSchema. The implicit solver calculates the internal or external force on unknown positions at the next time step, which shows better accuracy than the explicit solver by constructing a more complex linear equation. Afterwards, iterative CGLinearSolver is implemented to solve this equation. 
Each simulation is performed by specifying the solving iterations (without GUI visualisation). The grasping is considered to be successful if the position of the object remains within the threshold distance of the gripper after a certain elapsed time.

\subsection{Quality-diversity approach}

% This paper presents a method for the evolutionary design of diverse soft robotic grippers to enhance grasping success rates. While soft robotic design approaches have largely relied on bio-inspired and experience-driven methodologies, we propose a quality diversity (QD) approach to generate a collection of soft robotic grippers. By farming the design problem as a QD problem, our approach efficiently generates gripper designs with distinct features. To achieve this, we employ several quality diversity optimization algorithms, such as Novelty Search with Local Competition (NSLC), Multi-dimensional Archive of Phenotypic Elites (MAP-Elites) and Covariance Matrix Adaptation MAP-Elites (CMA-ME) to train gripper generators and compare the performance between those QD algorithms. The proposed method is applied to optimize the design of Fin Ray grippers, widely utilized in research and industry due to their flexibility, passive compliance, and durability. Our objective is to create a diverse collection of Fin Ray grippers with optimized grasping success rates and variations in weight and workspace. The obtained results demonstrate the feasibility and potential of our approach, enabling the generation of Fin Ray grippers capable of grasping various objects and offering a wide range of options for decision-makers.

The design updating is represented as a combinatorial optimisation problem. We generate a diverse collection of fin ray soft grippers through CMA-ME~\cite{fontaine2020covariance}, a quality-diversity algorithm combining the adaptation mechanisms of CMA-ES~\cite{DBLP:journals/ec/HansenO01} with the archiving mechanism of MAP-Elites~\cite{DBLP:journals/nature/CullyCTM15}. It specialises in continuous domains and is significantly more sample-efficient than other QD algorithms. The workspace and the volumn of grippers are set as the features in the algorithm. As present in Figure~\ref{fig:framework}, the algorithm initialises certain populations and evaluates those designs by checking the preset threshold. The algorithm is terminal until reach the computational budget, providing the gripper designs with rich features. All those optimised grippers demonstrate superior performance under the simulation environment.
% Several designs are finally selected to be manufactured with one-shot 3D printing for the experimental demonstration of the commercially available 7-DoF robotic manipulator.

    %  \begin{figure}[ht]
    %     \centering
    %     \includegraphics[width=1\linewidth]{Finray_framework.png}
    %     \caption{Overview of fin ray gripper designs generator}
    %     \label{fig:ss}
    % \end{figure}

    % \begin{figure}[ht]
    %         \centering
    %         \includegraphics[width=0.8\linewidth]{workspace.PNG}
    %         \caption{workspace}
    %         \label{fig:workspace}
    %     \end{figure}
    
Workspace is defined as the range of positions a robot can reach to interact with its physical environment. Thus, a robotic manipulator’s workspace consists only of all possible positions of the robot's tip, or \textit{robotic gripper}. Soft grippers can interact with target objects at all points within the volume swept due to their actuator deformations. All passive and active deformations possible for the soft gripper actuators contribute to the workspace. This swept volume or workspace volume strongly correlates with the payload size range and grasping versatility of the gripper~\cite{jain2023multimodal}. 
Here, the workspace of the soft gripper is varied when the fingers' relative distance changes. The maximum area of the equilateral triangle that the three fingers cover is calculated to represent the maximum workspace, which is treated as the first feature. 

    \begin{equation}
        fea_1 = \frac{3\sqrt{3}}{4} \times d_{mount}^2
    \end{equation}

% In addition, another feature that we are interested in building the archive of CAM-ME is the morphological complexity, which has drawn much attention from the soft robotics research area~\cite{corucci2018evolving,DBLP:conf/ecal/CluneL11,DBLP:conf/alife/0001A14}. 
% As in~\cite{DBLP:journals/ploscb/AuerbachB14}, morphological complexity is here quantified in information-theoretic terms as the entropy of curvature (here also referred to as shape entropy), which is computed on the undeformed mesh representing the robot. 
% We use the measure of morphological complexity from~\cite{DBLP:journals/ploscb/AuerbachB14,DBLP:conf/icip/PageKSRA03}. To compute the complexity metric for a given gripper, we first compute the angular deficit $\Phi_j$ for each vertex $j$:

% \begin{equation}
%     \Phi_j = 2\pi - \sum_{i}\phi_i,
% \end{equation}

% where $\phi_i$ is the internal angle of each triangle $i$ where it meets vertex $j$. The deficit values are placed in a histogram over the range $[-2\pi, 2\pi)$ with bin width $\Delta$, which is normalized as a probability density function such that each bin $b$ contains a probability $p(\Phi_b)$. The morphological complexity is then equivalent to the entropy of the probability density function as follows,

In addition, the other feature we are interested in is the structure complexity, which can be represented using the total volume of the fin ray design. 
   
    % \begin{equation}
    %     fea_2 = V_{ribs}(N,t_{rib}) + V_{side}(H,L,W,L_{angle},L_{tip}, t_{flex},t_{back}) \\
    %     + V_{mount}(d_{mount}).
    % \end{equation}

   \begin{equation}
    \begin{split}
    fea_2 &= V_{mount}(d_{mount}) + V_{ribs}(N,t_{rib}) \\
    &\quad + V_{side}(H,L,W,L_{angle},L_{tip}, t_{flex},t_{back})
    \end{split}
    \end{equation}

% we evaluating the fin ray soft grippers via the Pyribs library~\cite{10.1145/3583131.3590374}, a QD optimization library maintained by the authors of CMA-ME. 

% Experimental pick and place task with 6 DoF robotic arms and considering grasping uncertainty.
% Maybe provide a benchmark experimental setup/database to evaluate the success rate for soft grasping as well. The success rate is defined differently: pick and place, contact force, vertical lift, etc. 

\section{Results and Discussion} \label{results}
% Experimental grasping tests across multiple objects to further evaluate the grasping rate of top-ranking designs. 
% Generate top-ranking designs with different features from QD.
% Video demonstration of grasping multiple objects.

% 1. sim tuning: tensile tests, mesh tuning (finger and object), speed and results

% 2. optimisation results with QD

% 3. experimental setup and results

    % \xing{fenjie}
    
To retrieve the high-fidelity SOFA result, material properties must be obtained experimentally. We designed the dogbone elastomer samples and performed the universal tensile test based on the ASTM D412 standard. An Instron 34SC-5 equipment was used to collect the experimental tensile results 5 times, and the average strain-stress curve was plotted to fit the linear elastic material model in SOFA. Several digitally mixed materials made from the Aglius 30 and Vero family were tested, and the one with a shore hardness of 85A was eventually implemented in the simulation tests. The material was assigned with an elastic modulus of 11.6 MPa and passion ratio of 0.49 due to the incompressible nature of rubber-like soft material. 

The same quality of Tetrahedron mesh has been applied across all fingers, while the mesh for the objects was set to be more coarse to prevent penetration from the master nodes (rigid objects) to slave nodes (soft fingers). The minimum and maximum mesh range was set to be [0.5, 1] and [2, 4] using Gmsh API, respectively. Modifying the mesh quality can allow a different fidelity of simulation results as well, which has been tested before the implementation of the current mesh quality.
For the iterative linear solver, we set up a maximum iteration of 1000 and a tolerance of 1e-6. The residual of the linear matrix is reduced each time with an increase in iteration.

\begin{figure}[ht]
                \centering
                \includegraphics[width=0.99\linewidth]{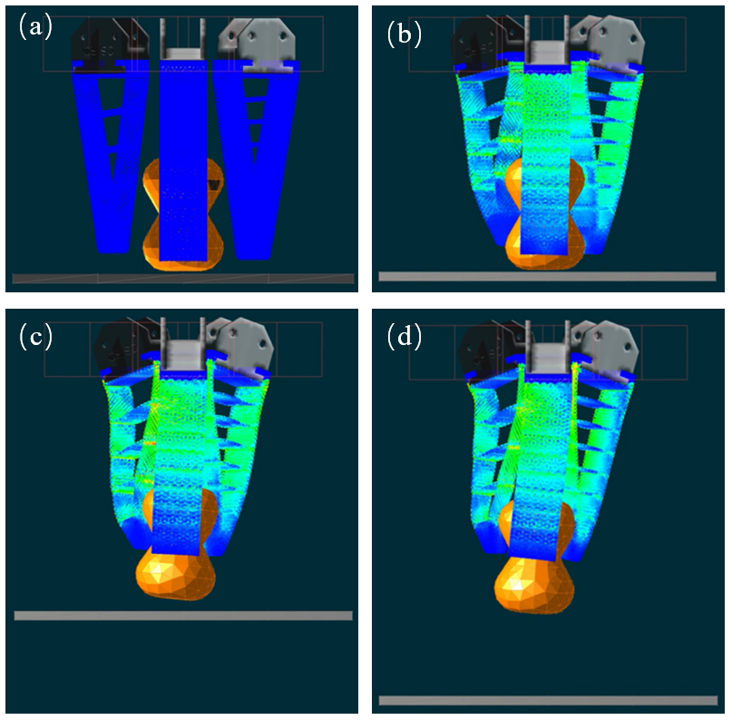}
                \caption{SOFA simulation for a complete grasping process. The grasping is treated as a success if the object is still within the preset height in the scene.}
                \label{fig:sim}
            \end{figure}

A complete SOFA simulation process is demonstrated in Figure \ref{fig:sim} and \textbf{Supplementary Video SV2}. 
    
\begin{figure}[h]
\centering
\begin{subfigure}[b]{0.22\textwidth}
   \includegraphics[width=0.95\linewidth]{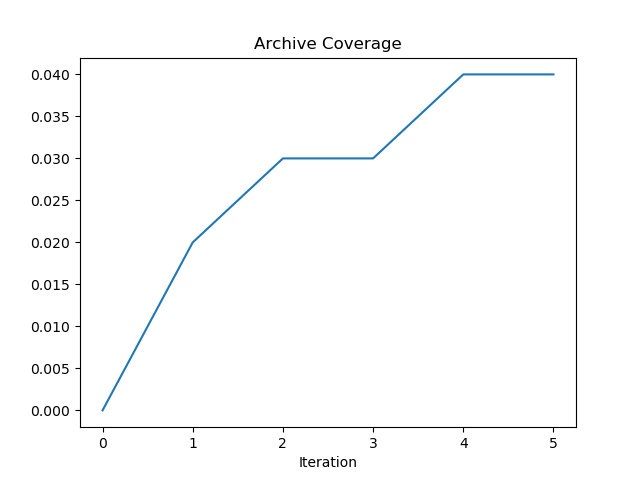}
  \caption{\textit{Archive coverage trade through the evaluation}}
  \label{fig:coverage}
\end{subfigure}
\begin{subfigure}[b]{0.22\textwidth}
   \includegraphics[width=0.9\linewidth]{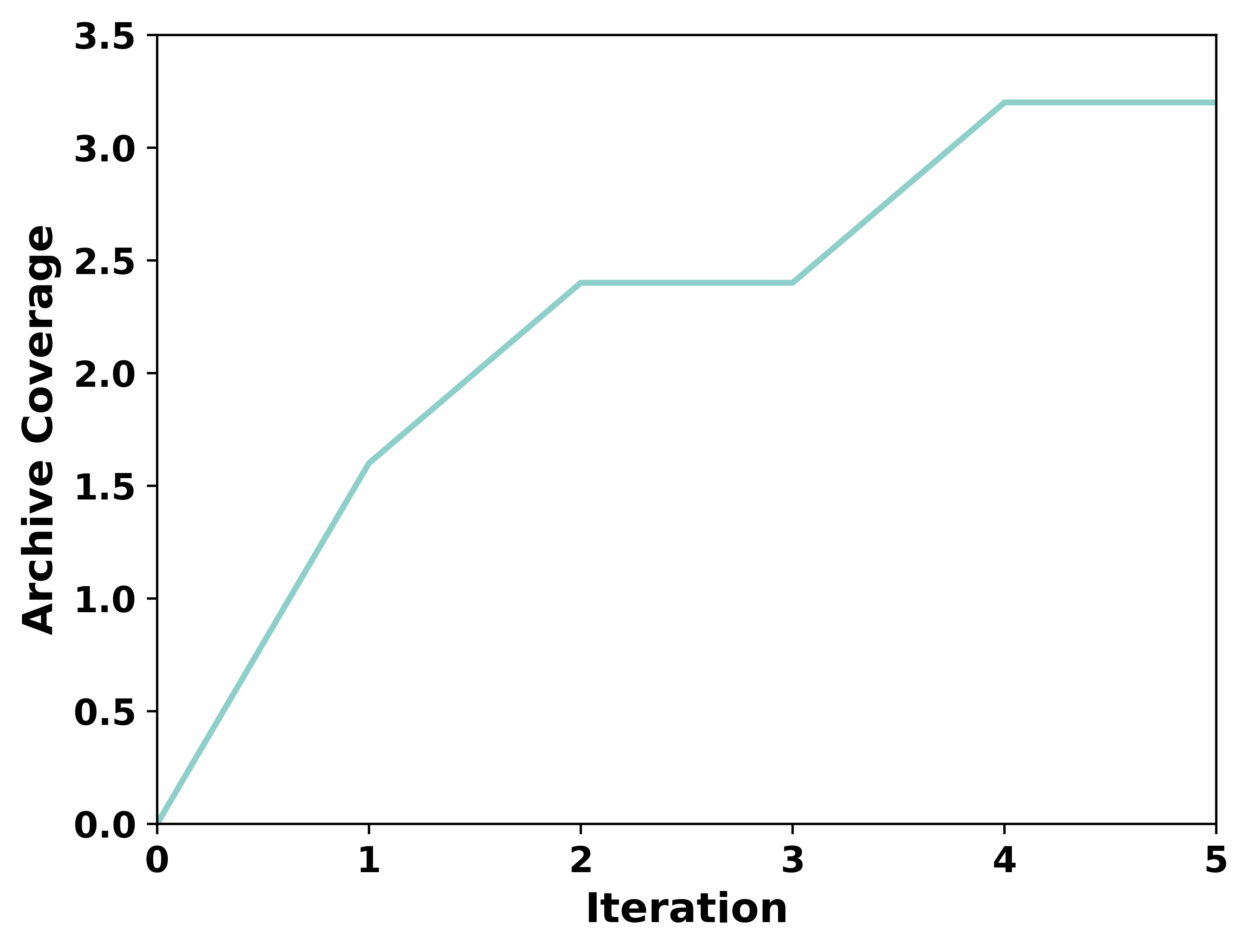}
  \caption{\textit{QD score through the evaluation} }
  \label{fig:qdScore}
\end{subfigure}
\caption{Evaluate indicates of the Quality diversity algorithm}
\label{fig:cov_qd}
\end{figure}

The experiment is run on a Dell Precision Workstation with Inter I9-12900H CPU and is configured to terminal after $5$ iterations considering time complexity with setting batch size to $15$. 
% GPUCUDA calculation and parallel computing for SOFA are feasible to speed up the simulation. 
Figure~\ref{fig:cov_qd} presents the improvement trade of the quality diversity algorithm via archive coverage and QD score, where archive coverage presents how well the archive represents the diversity and quality of the solutions found so far and QD score~\cite{tjanakaquantifying} is a holistic metric which sums the objective values of all cells in the archive. It can be seen that even in $5$ generations, the coverage of the archive as well as the quality of the resulting design grows.

\begin{figure}
    \centering
    \includegraphics[width=0.8\linewidth]{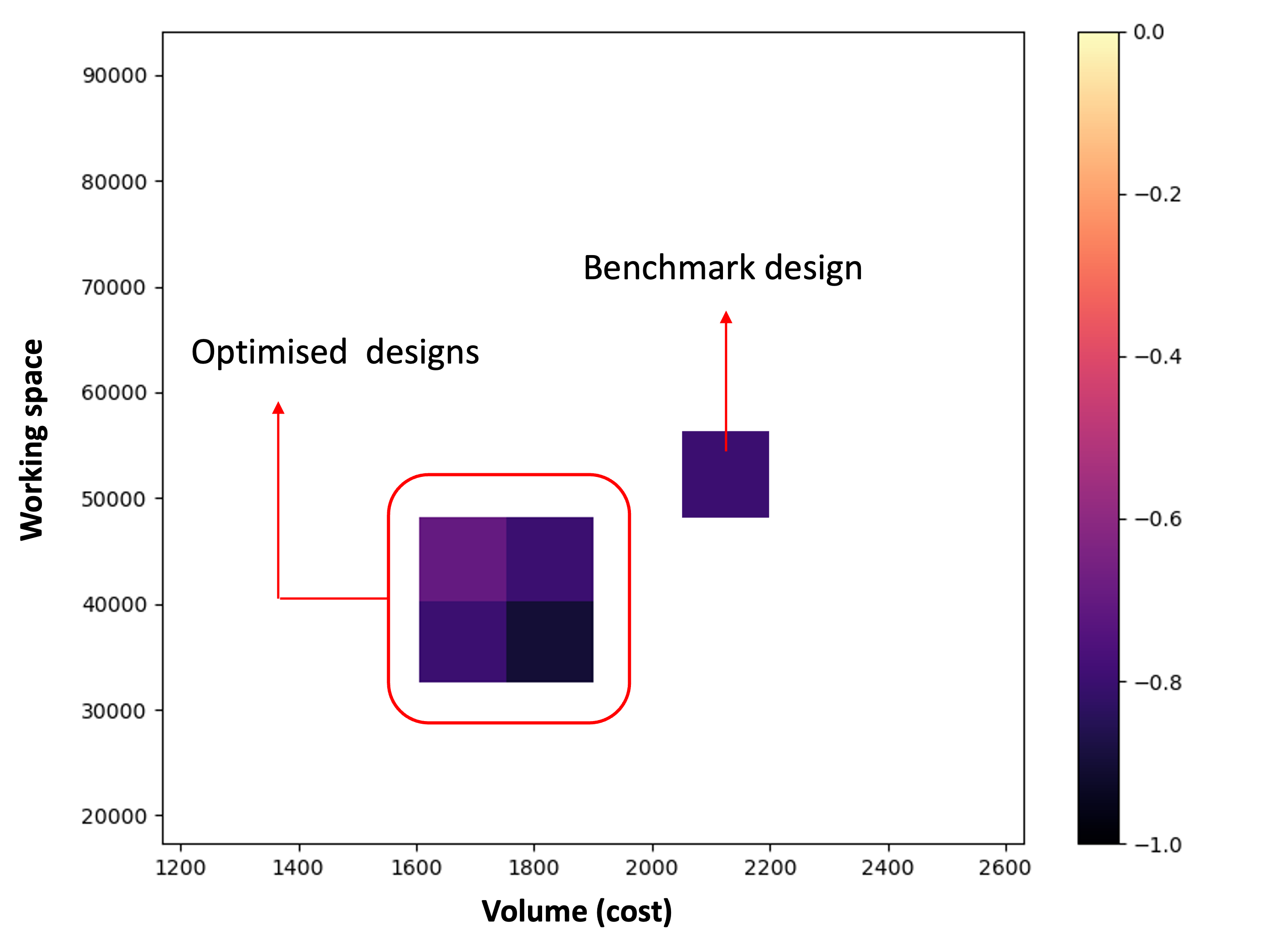}
    \caption{Distribution of high-performance designs and benchmark design}
    \label{fig:heat}
\end{figure}

Figure~\ref{fig:heat} presents the distribution of high-quality designs of the final archive within the behavioural space and the corresponding cell in the behaviour space of the bench gripper. The presented heatmap clearly illustrates that the proposed framework can provide a diverse set of gripper designs. Considering volume (cost), the designs obtained by the framework perform better than the benchmark and show competitive performance in the workspace. 

\begin{figure*}
         \centering
        \includegraphics[width=0.95\linewidth]{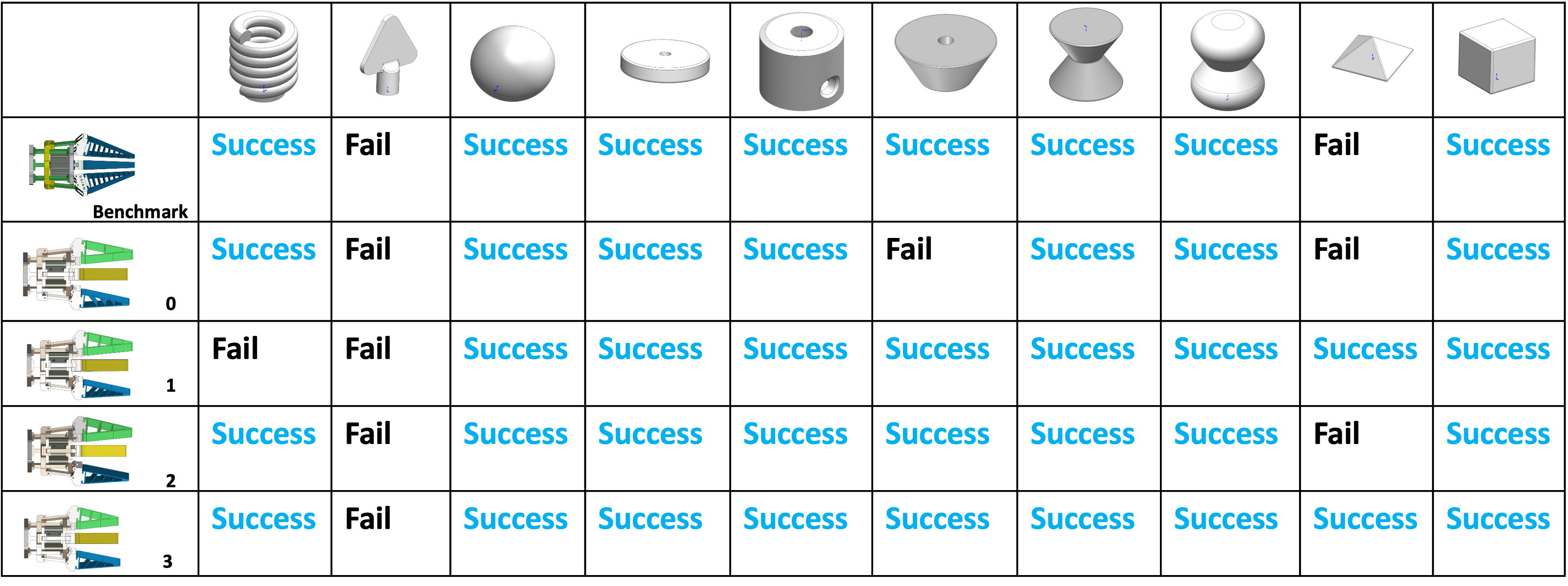}
        % \caption{larx}
    \caption{Comparison of the grasping performance}
        \label{fig:final}
    \end{figure*}

Moreover, Figure~\ref{fig:final} presents the success or failure of the benchmark design and optimised designs respectfully. It can be observed from the figure that all the grippers fail to pick object $2$, and designs $2$ and $3$ perform the same as the benchmark design with a lower cost. All grippers perform well in grasping simple primitive objects and achieve a grasping success rate of no less than 0.7.

% \begin{table*}[ht]
% \centering
% \begin{tabularx}{\textwidth}{|X|X|X|X|XXXX|}

% \cline{1-8}
% Category& Name & Index & Benchmark & Solution1 & Solution2 & Solution3 & Solution4 \\
% \hline

% \multirow{10}{*}{Objects} & 1 &Pasta & 1 &1&0&1&1\\
% % \cline{2-8}
% & 2&Triangle    &   0& 0& 0& 0& 0\\
% % \cline{2-8}
% & 3&Sphere & 1 & 1 & 1& 1& 1 \\
% % \cline{2-8}
% & 4&Disk &  1 &1 &1 &1 & 1\\
% % \cline{2-8}
% & 5&Cylinder& 1  &  1 & 1&1& 1\\
% % \cline{2-8}
% & 6&Flan&1&0&1&1& 1\\
% % \cline{2-8}
% & 7&Diablo &1&1&1&1& 1\\
% % \cline{2-8}
% & 8&Hourglass&1&1&1&1&1 \\
% % \cline{2-8}
% & 9 &Diamond &0&0&1&0&1\\
% % \cline{2-8}
% & 10 &Cube &1&1&1&1&1 \\
% % \cline{2-8}

% \hline
% \multirow{3}{*}{Performance} & \multicolumn{2}{c|}{Fitness} & 0.8 & 0.7  & 0.8  &  0.8 &  0.9 \\
% % \cline{2-8}
%  &  \multicolumn{2}{c|}{Feature1} & 2078.46 &1672.25 & 1693.06 & 1756.33 &  1754.62\\
% % \cline{2-8}
%  &  \multicolumn{2}{c|}{Feature2} & 51139.59 &41063.59  & 39522.40  &   42406.26& 39912.05  \\
% \hline
% \end{tabularx}
% \caption{Comparison of the grasping performance, while 0 and 1 represent a fail and success grasp, respectively}
% \end{table*}

\section{Conclusions and future work} \label{conclusions}

This paper presents a computational design framework for exploring the evolution of soft robots in contact-based grasping tasks. The study focuses on finger-based soft robotic grippers and aims to bridge the gap in the current understanding of the design space and performance evaluation. The proposed framework introduces a highly automated design optimization process that utilizes a quality-diversity optimization algorithm which allows for the generation of gripper diversity by exploring a significantly higher dimensional design space. To evaluate the performance of the designed grippers, a rapid and high-fidelity contact-based simulation is proposed within the SOFA framework. This simulation accurately captures the complete soft grasping motion and measures the grasping success rate for performance evaluation. 
The framework also incorporates a comprehensive evaluation of the automated design optimization process using both simulation experiments. The experimental discussion demonstrates the potential of evolutionary techniques in producing diverse and high-performance soft robotic grippers. The proposed framework can serve as a valuable tool for researchers and designers in the field of soft robotics, enabling them to explore and optimize complex design spaces for various applications.

Future work may explore the effect of finger numbers and more irregular configurations on more complex object datasets. The dataset covers geometrically symmetric target objects, which makes it easier for such grippers that have symmetrical arrangements in the gripper mount. More challenges are faced for objects with irregular shapes or different material properties (mainly soft materials).  Additionally, it is worth investigating further improving the fidelity of the simulation model by employing the sim-to-real transfer algorithms and conducting the physical experiments to verify the real grasping performance.

\section{Acknowledgement}
We acknowledge Lois Liow's support in designing the object database.

\bibliographystyle{IEEEtran}
\bibliography{IEEEabrv,bib}

\end{document}